\documentclass{article}

\usepackage{arxiv}
\usepackage{amsmath, amsfonts, amssymb}
\usepackage{graphicx}
\usepackage{hyperref}
\usepackage{booktabs}
\usepackage{float}
\usepackage{tikz}
\usetikzlibrary{arrows.meta, positioning, shapes.geometric}
\makeatletter
\def\headeright{}
\def\undertitle{}
\makeatother
\AtBeginDocument{
  \setlength{\headheight}{23pt}
  \addtolength{\topmargin}{-9pt}
}

\title{Intermediate Artifacts as First-Class Citizens:\\
A Data Model for Durable Intermediate Artifacts in Agentic Systems}

\author{
  Josh Rosen \\
  ThruWire, Inc.
  \And
  Seth Rosen \\
  ThruWire, Inc.
}

\begin{document}

\maketitle

\begin{abstract}
Many AI systems are organized around loops in which models reason, call tools, observe results, and continue until a task is complete. These systems often produce final artifacts such as memos, plans, recommendations, and analyses, while the intermediate work that shaped those outputs remains ephemeral. For multi-step, revisable AI work, final artifacts are often lossy projections over upstream state.

We argue that such systems should preserve durable, inspectable intermediate artifacts: typed, structured, addressable, versioned, dependency-aware, authoritative, and consumable by downstream computation. These artifacts are not the model's private chain-of-thought. They are maintained work products such as evidence maps, claim structures, criteria, assumptions, plans, transformation rules, synthesis procedures, unresolved tensions, and partial products that later humans and agents can inspect, revise, supersede, and improve.

The contribution is a systems-level data model. We distinguish intermediate artifacts from chat transcripts, memory, hidden chain-of-thought, narration, thinking, and final answers; formalize additive and superseding update semantics with explicit current-state resolution; describe how artifact lineage supports durable intermediate state across revisions; and argue that evaluation must target maintained-state quality, not only final-output quality. The claim is not that artifacts make models smarter. It is that durable intermediate artifacts make AI-generated work more inspectable, revisable, and maintainable over time.
\end{abstract}

\section{Introduction}

Many current AI systems are organized around iterative loops in which a model reasons, calls tools, observes results, and continues until it decides it is done. Such loops are powerful for one-off tasks. This paper is about what should be preserved when that work must later be inspected, revised, and extended.

Most current interfaces still privilege final answers. Some expose narrated thinking or chain-of-thought-like text, even though such narration may be post hoc or otherwise unfaithful to the model's internal process. Other state may exist only in hidden model computation, controller logic, or transcript context. None of these is a reliable maintained-state boundary. Hidden state is not inspectable. Transcript state is readable but weakly structured and difficult to revise locally.

A multi-step process that produces an evidence digest, claim matrix, criteria artifact, plan, and final memo is therefore not well modeled as one prompt plus one answer. The final memo is a rendering over upstream work. If that upstream work disappears, later revision is forced to reconstruct the very state it should have been able to inspect directly.

This paper argues that revisable AI work should preserve durable, inspectable intermediate artifacts. We assume an execution substrate capable of staged execution, dependency tracking, and replay under change \cite{rosen2026executionlineage}. This paper focuses on the semantics of maintained intermediate artifacts within such a substrate: what kinds of intermediate objects should remain inspectable, how they are treated as current, and how they are superseded and improved over time.

The thesis is:
\begin{quote}
In LLM-mediated work, the relevant durable state is not only the final output but the maintained intermediate artifacts from which that output was produced. Those artifacts should remain addressable, authority-bearing, supersession-aware, and consumable inside the substrate so that later work can improve them at the source.
\end{quote}

The paper makes four contributions. First, it formulates an artifact-centric data model for revisable LLM systems and defines the properties that make an artifact first-class, including authority and current-state resolution. Second, it distinguishes \emph{artifact lineage} from \emph{prompt lineage}, memory, and final outputs. Third, it introduces additive versus superseding update semantics as a mechanism for maintaining coherent intermediate state over time. Fourth, it argues that evaluation must target maintained-state quality and intervention surfaces, not only final-output acceptability.

\section{Motivating Example}

Consider a policy-analysis system that produces a recommendation memo about telehealth expansion. A plausible execution includes source digests for utilization, reimbursement, operations, and access-cost tradeoffs; a claim matrix grounded in those sources; a tension analysis that preserves unresolved conflicts; a recommendation-criteria artifact; an implementation plan; and finally a memo.

Now suppose a year-one budget-neutrality constraint arrives after an initial draft. The important question is not only whether the final memo should change. It is what should regenerate.

The relevant intervention point is the recommendation-criteria artifact. That artifact should be revised so that year-one budget neutrality becomes a hard constraint. Once that criteria artifact changes, the system should regenerate the downstream intermediate artifacts that depend on it. The implementation plan should be rebuilt because rollout sequencing now depends on the new constraint. The recommendation synthesis should be rebuilt because the option-ranking logic has changed. The final memo should then be rebuilt from those regenerated intermediates. Other upstream artifacts, such as utilization evidence or tensions that do not depend on the changed criteria, should remain untouched.

The key point is that the relevant state is not only the final memo. The maintained body of work is the intermediate layer from which later revisions should proceed.

\begin{figure}[H]
\centering
\begin{tikzpicture}[
  font=\small,
  scale=0.88,
  transform shape,
  node distance=7mm and 10mm,
  box/.style={draw, rounded corners, align=center, minimum width=29mm, minimum height=8mm},
  artifact/.style={draw, rounded corners, align=center, minimum width=30mm, minimum height=8mm, fill=black!4},
  state/.style={draw, rounded corners, align=left, minimum width=34mm, minimum height=8mm, fill=black!2},
  arrow/.style={-{Stealth[length=2.2mm]}, thick},
  dashedarrow/.style={-{Stealth[length=2.2mm]}, dashed, thick}
]
\node[font=\bfseries] at (0,0) {Final-answer loop};
\node[box] (goal) at (0,-0.9) {Task / update request};
\node[box] (transcript) at (0,-2.2) {Prompt assembly\\transcript + notes};
\node[box] (loop) at (0,-3.5) {LLM / tool loop};
\node[box] (answer) at (0,-4.8) {Revised final answer};
\node[state] (hidden) at (3.8,-3.5) {intermediate work is\\summarized, overwritten,\\or kept implicit};
\draw[arrow] (goal) -- (transcript);
\draw[arrow] (transcript) -- (loop);
\draw[arrow] (loop) -- (answer);
\draw[dashedarrow] (hidden.west) -- (loop.east);

\node[font=\bfseries] at (10.5,0) {Artifact-first system};
\node[artifact] (src) at (10.5,-0.9) {Source and context artifacts};
\node[artifact] (claims) at (8.2,-2.2) {Claim matrix};
\node[artifact] (tensions) at (12.8,-2.2) {Tension analysis};
\node[artifact] (criteria1) at (10.5,-3.5) {Criteria v1};
\node[artifact] (criteria2) at (10.5,-4.8) {Criteria v2\\(supersedes v1)};
\node[artifact] (plan) at (10.5,-6.1) {Implementation plan};
\node[artifact] (memo) at (10.5,-7.4) {Final memo};
\draw[arrow] (src.south west) -- (claims.north);
\draw[arrow] (src.south east) -- (tensions.north);
\draw[arrow] (claims.south east) -- (criteria1.north west);
\draw[arrow] (tensions.south west) -- (criteria1.north east);
\draw[dashedarrow] (criteria1.south) -- (criteria2.north);
\draw[arrow] (criteria2.south) -- (plan.north);
\draw[arrow] (plan.south) -- (memo.north);
\node[state] at (15.6,-4.8) {downstream nodes\\consume typed artifacts\\with lineage and versioning};
\end{tikzpicture}
\caption{A final-answer loop collapses intermediate state into a transcript and answer. An artifact-first system preserves typed intermediate artifacts, allows explicit supersession, and keeps downstream dependencies addressable.}
\label{fig:artifact-model}
\end{figure}
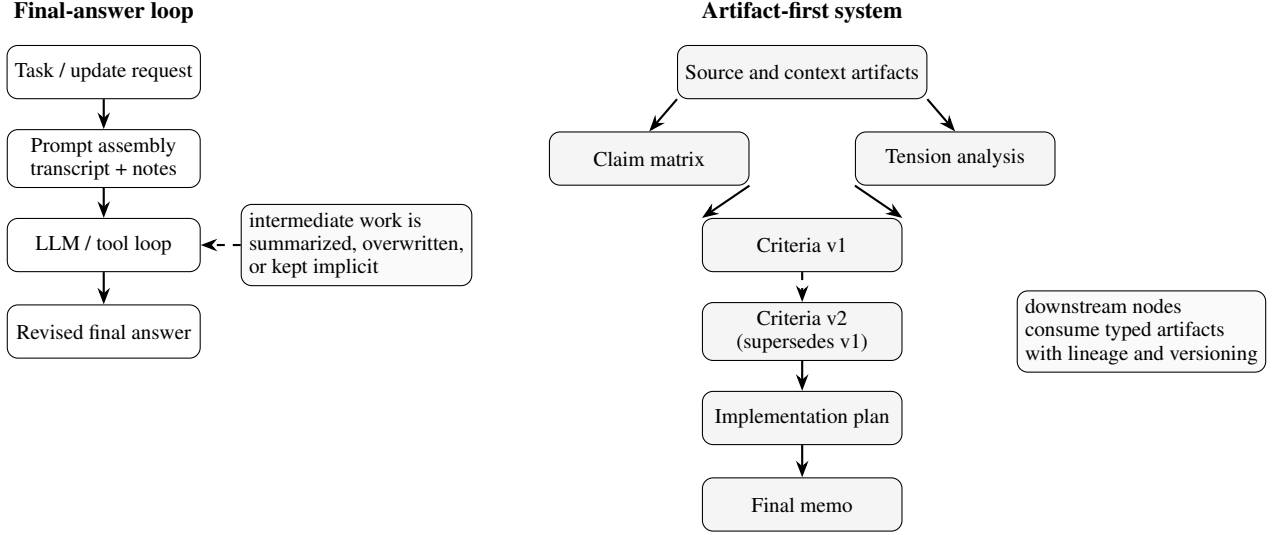

Figure~\ref{fig:artifact-model} contrasts these two views. On the left, a loop collapses intermediate work into an answer-oriented transcript. On the right, an artifact-first system preserves typed intermediate state, supports revision via explicit supersession, and makes downstream dependencies addressable.

\section{Background and Related Work}

\subsection{Agent Loops, Tool Use, and Intermediate Reasoning}

Agentic LLM systems commonly interleave reasoning, tool use, and revision inside iterative loops. ReAct \cite{yao2023react} remains the canonical example of this paradigm, while Toolformer \cite{schick2023toolformer}, AutoGen \cite{wu2023autogen}, CAMEL \cite{li2023camel}, Voyager \cite{wang2023voyager}, and MetaGPT \cite{hong2023metagpt} explore richer tool ecosystems, role structures, and multi-agent coordination. These systems demonstrate that substantial capability gains can be achieved by placing models inside structured harnesses rather than treating them as isolated text generators.

Work on chain-of-thought, scratchpads, self-refinement, and search further shows that intermediate reasoning can matter operationally \cite{nye2021scratchpads,wei2022cot,wang2022selfconsistency,zhou2022leasttomost,yao2023tree,shinn2023reflexion,madaan2023selfrefine,zhou2023lats}. This literature is directly relevant because it establishes the value of intermediate reasoning structures. At the same time, it generally treats such intermediates as execution-local support for a run rather than durable, addressable state for future revision. It is also important not to equate surfaced chain-of-thought narration with faithful access to the model's actual internal process: Turpin et al. \cite{turpin2023unfaithful} show that chain-of-thought explanations can be plausible yet misleading and can omit features that actually drove the prediction. Our claim is not that raw chain-of-thought should be persisted. It is that durable reasoning structures should exist as maintainable artifacts in the substrate.

\subsection{Harnesses, Memory, and Execution Substrates}

Recent work increasingly treats the agent harness itself as an object of study. Natural-Language Agent Harnesses \cite{pan2026nlah} and modular harness research \cite{zhang2025harness} argue that important behavior lives in controller logic, memory, and interfaces around the model. Memory-oriented systems such as Agent Workflow Memory \cite{wang2024awm}, WorkflowLLM \cite{fan2024workflowllm}, LEGOMem \cite{han2025legomem}, Memory-R1 \cite{yan2025memoryr1}, and Memori \cite{borro2026memori} likewise recognize that sustained agentic work depends on persistent state and retrieval. In the framing of this paper, the substrate includes both the harness that executes steps and the maintained artifacts that those steps generate. Memory asks what should be retained and later retrieved. This paper asks what intermediate work should remain alive as maintained state inside that substrate, how it becomes current, and how later revisions supersede it.

\subsection{Dependency Graphs, Lineage, and Provenance}

Dependency-graph and data systems have long relied on explicit dependencies, materialized intermediates, and lineage-aware recomputation \cite{isard2007dryad,zaharia2016spark,airflowdocs,dbtdocs}. The same general lesson appears in data provenance research \cite{buneman2001why,freire2006vistrails} and in provenance-aware LLM-agent work \cite{souza2025provenance}: once work becomes staged and collaborative, hidden dependencies and ephemeral outputs become liabilities.

Our proposal differs in target object. Classical provenance systems usually record how deterministic data products were produced. Agentic LLM systems add stochastic generation, prompt-conditioned steps, and richer intermediate semantics. The need for lineage remains, but the primary durable object here is not only tables, files, or final rendered outputs. It is intermediate analytical state intended for both humans and downstream model steps. Provenance records how an output was produced; artifact lineage also specifies which intermediate state is current, superseded, reusable, or authoritative for future work.

\subsection{Notebooks, Literate Workflows, and Intermediate Representations}

The importance of preserving intermediate work is not unique to AI systems. Literate programming \cite{knuth1984literate} and notebooks \cite{kluyver2016jupyter} both treat intermediate computational state as a meaningful object of inspection and revision rather than a hidden implementation detail. Similarly, compiler design depends on explicit intermediate representations because they create analyzable, transformable, and optimizable boundaries between stages \cite{lattner2021mlir}. These traditions matter here because they show the value of explicit intermediate forms. Agentic LLM systems, however, require additional semantics for stochastic generation, prose artifacts, supersession, published reasoning structures, and maintained artifact state.

\subsection{Why Knowledge Graphs and Wikis Are Not Enough}

Knowledge graphs and wikis preserve facts and relations, and they are valuable long-term memory surfaces. They do not usually provide the full combination of execution identity, procedural dependencies, invalidation semantics, and active-state resolution needed here by default. Knowledge-graph systems usually capture static semantic relationships rather than execution identity, procedural dependencies, or invalidation semantics \cite{hogan2021knowledgegraphs}. Wikis can store notes, but they do not ordinarily specify which downstream outputs should be recomputed when an upstream artifact changes. A linked knowledge base can tell the system that one artifact refers to another; it does not by itself specify the step structure by which intermediate artifacts are generated, combined, and regenerated when upstream state changes.

Taken together, prior systems preserve traces, memories, workflows, provenance, or knowledge structures. The missing abstraction is maintained intermediate artifact state inside a harness-backed work substrate: typed objects with currentness, authority scope, supersession semantics, and downstream consumability. Provenance centers production history. Workflow engines center execution order. Notebooks center editable sessions. Versioned documents center successive renderings. Memory systems center retention and retrieval. What is new here is the combination of semantics around a single object: a durable intermediate artifact that is step-produced, inspectable, authority-scoped, supersession-aware, dependency-addressable, and directly consumable by later humans and model-mediated steps.

Table~\ref{tab:surface-comparison} summarizes the distinction between adjacent state surfaces.

\begin{table}[H]
\centering
\small
\begin{tabular}{p{0.18\linewidth}p{0.21\linewidth}p{0.22\linewidth}p{0.27\linewidth}}
\toprule
Surface & Stores & Main question & Limitation \\
\midrule
Transcript & Interaction history & What happened? & Poor computational boundary \\
Chain-of-thought / scratchpad & Execution-local reasoning trace & How did the model reason locally? & Usually private, transient, and non-authoritative \\
Memory & Retrieved prior context & What should be remembered? & Weak authority and revision semantics \\
Provenance & Production history & How was this produced? & Does not necessarily define current state \\
Knowledge graph / wiki & Facts and relations & What is known? & Weak procedural lineage and invalidation semantics \\
Intermediate artifact model & Maintained artifact state & What exists now, what replaced what, and what depends on it? & Requires schema and substrate discipline \\
\bottomrule
\end{tabular}
\caption{Adjacent state surfaces answer different questions. The artifact model is aimed at maintained artifact state rather than only history, retrieval, or factual storage.}
\label{tab:surface-comparison}
\end{table}

\section{Problem Formulation}

We focus on systems whose outputs are revised over time rather than consumed only once. The central question is what state such a system leaves behind so that later work can inspect, reuse, update, and recompute it coherently.

Transcript-centric systems often answer: persist the conversation history, perhaps summaries, perhaps some cross-conversation memory, and perhaps the final answer. This does not usually provide a structured intermediate object that downstream consumers can address directly. Revision therefore operates on rendered surfaces rather than upstream causes.

The same gap appears in a pure knowledge-base approach. A system can store facts, notes, and links between intermediate objects without preserving how the substrate generated those objects from one another. But once step structure and dependency semantics are lost, a change to one intermediate artifact does not tell the system what should be regenerated, recombined, or superseded downstream.

The underlying issue is a mismatch between staged work and maintained state. A system may decompose work into explicit boundaries, dependencies, and intermediate products. But if the outputs of those boundaries are flattened back into prompt context, or preserved only as linked notes without regeneration semantics, the structure exists only weakly after execution.

We use reasoning-relevant artifacts as a motivating case, but the underlying systems property is broader: maintained intermediate state. In a decomposed system, selected outputs at intermediate state boundaries should be represented as durable artifacts with explicit semantics for identity, dependency, authority, and revision. Chain-of-thought is the model's transient internal reasoning trace; intermediate artifacts are the system's durable maintained state.

\subsection{Final Artifacts as Lossy Projections}

The problem is not that a final artifact contains no information about its upstream reasoning. The problem is that it contains that information in compressed, entangled, and non-authoritative form. A reconstructed rationale is not the same as preserved lineage.

If intermediate artifacts are not persisted, later revision is forced to operate on a flattened projection. This makes improvement harder to localize and easier to drift.

\subsection{Maintained-State Degradation}

These failures can be grouped under maintained-state degradation. Three recurring forms are state flattening, layered revision debt, and reasoning provenance loss.

\paragraph{State flattening.} Upstream intermediate state is compressed into a final output without preserving the intermediate objects needed to inspect, revise, or reuse it.

\paragraph{Layered revision debt.} Systems repeatedly patch rendered outputs while leaving underlying intermediate state unavailable, stale, or unauthoritative.

\paragraph{Reasoning provenance loss.} Useful decompositions, criteria, transformations, decision procedures, reconstruction methods, or tradeoff analyses never enter maintained substrate state and therefore cannot be built upon in later revisions.

These failures may remain invisible when the current final artifact looks acceptable; they matter because they degrade the state surface on which future improvement depends.

\section{Artifact-Centric Data Model}

\subsection{First-Class Artifact Properties}

An artifact-centric system treats intermediate outputs as first-class state when they satisfy seven properties.

\paragraph{Typed.} An artifact has an explicit role or schema surface. The type may be lightweight, such as a structured memo section or decision matrix, or more rigid, such as a JSON payload. The point is not maximal formality; it is that downstream consumers should know what kind of object they are reading.

\paragraph{Structured.} The artifact has internal organization beyond arbitrary free text. Structure may be tabular, sectioned, key-value, XML-like, or otherwise canonicalized.

\paragraph{Addressable.} The artifact can be referred to directly, rather than only by replaying the conversation that happened to produce it. Addressability is the precondition for reuse, inspection, and comparison across runs.

\paragraph{Versioned.} The system can distinguish an earlier artifact from a later one. Versioning does not require user-facing version numbers in every case; it requires identity surfaces that let the system know when an artifact is the same, new, or revised.

\paragraph{Dependency-aware.} The artifact records which upstream state it depends on. This is what allows downstream state maintenance to be principled rather than heuristic.

\paragraph{Authoritative.} The system can determine whether an artifact is active current state, historical state, superseded state, or an alternative branch. This matters because downstream consumers need to know not only that an artifact exists, but whether it is the current artifact they should consume.

\paragraph{Consumable downstream.} The artifact is not persisted only for audit. It is an operational dependency surface for other nodes or steps.

Together these properties define a first-class artifact as more than a saved message. It is a typed unit of maintained artifact state with an explicit currentness boundary.

\subsection{Core Record Semantics}

At the data-model level, a minimal artifact record needs fewer fields than a full implementation. The mandatory semantics are:
\begin{itemize}
\item \textbf{artifact identity:} a stable address for the committed artifact instance
\item \textbf{artifact family:} the conceptual object across revisions, such as ``current recommendation criteria''
\item \textbf{role:} the functional purpose of the artifact, such as evidence digest, claim matrix, criteria, or plan
\item \textbf{scope:} the authority boundary within which active-state resolution occurs
\item \textbf{status:} whether the artifact is active, superseded, historical, or an active branch alternative
\item \textbf{dependencies:} the upstream artifacts materially consumed to produce it
\item \textbf{lineage:} the execution boundary that produced it and the supersession relations that connect it to prior artifacts
\item \textbf{payload:} the structured content that later humans and systems inspect or consume
\end{itemize}

Other fields are implementation choices rather than core semantics: timestamps, hashes, storage references, block identifiers, caching metadata, streaming identifiers, display names, or transport-layer envelopes.

\paragraph{Artifact family.} A family groups revisions of the same conceptual object over time. For example, \texttt{criteria:v1} and \texttt{criteria:v2} may belong to the same family even though only one is currently active.

\paragraph{Role.} A role names what the artifact is for in downstream work. It is not merely a file type. Two JSON payloads may have different roles if one is a claim matrix and the other is an implementation plan.

\paragraph{Scope.} A scope defines the authority boundary within which currentness is resolved. It may correspond to a project, branch, scenario, customer, jurisdiction, or other explicitly modeled context. Scope answers the question ``current for whom or for what boundary?''

\paragraph{Status.} Status records whether the artifact should be treated as active, historical, superseded, or as one active branch among several permitted alternatives.

\paragraph{Dependency and lineage semantics.} Dependencies record the upstream maintained artifacts materially consumed by this artifact. Lineage records both production provenance and supersession relations. Together they determine what can be trusted as current and what should be recomputed when a revision occurs.

\subsection{Local Versus Maintained Artifacts}

An artifact-centric model distinguishes local working artifacts from maintained artifacts in the substrate, and it does not treat the final answer as the only durable object. In staged or graph-executed systems, work is organized into execution steps that generate intermediate artifacts during reasoning, tool use, decomposition, synthesis, or revision. Step-level and block-level artifacts may both be part of the substrate. The distinction is instead between artifacts that remain local working state inside a step and artifacts that are kept as maintained, dependency-addressable state that later steps, humans, or agents can continue to inspect and improve.

This distinction matters because maintained artifacts define the durable interface contract between boundaries. Downstream consumers use them as typed state inputs under resolved identities rather than reconstructing state from transcript context. Local working artifacts may still be persisted for inspection and revision, but they are not automatically authoritative for downstream use simply because they were produced during work.

\subsection{Deliberate Substrate Entry}

First-class artifacts should become maintained state deliberately. A system may internally generate candidate drafts, narration, or local working traces, but only some artifacts become maintained state that the system expects others to keep iterating on. This distinction matters. It distinguishes a durable artifact from raw private reasoning. It also separates two concerns that are often conflated:

\begin{itemize}
\item \textbf{How the model reasoned locally}
\item \textbf{What state the system chose to maintain in the substrate}
\end{itemize}

The first may remain private, partial, or non-canonical. The second must be reviewable and fit for downstream use.

Artifact payloads need not be final prose. They may encode evidence maps, claim graphs, criteria, assumptions, plans, partial products, decision procedures, transformation rules, reconstruction methods, unresolved tensions, or other agent-accessible structures. Reasoning structures are one important class of artifact payload, not the entire artifact model.

Operationally, this is a maintained-state property rather than a disclosure policy. The question is not whether a model exposed chain-of-thought text to a user. It is which artifacts the system keeps as maintained state, under what identity, with what authority scope, and with what supersession semantics.

\section{Artifact Lineage and Supersession Semantics}

\subsection{Prompt, Execution, and Artifact Lineage}

Prior work on execution lineage established the lineage of computation that produced an output under explicit graph execution. For the present paper, it is useful to distinguish three related but non-identical kinds of lineage.

\paragraph{Prompt lineage.} Prompt lineage records what text, instructions, and transcript state led to a model response. It is useful for audit and debugging, but it is not a stable substrate for composition.

\paragraph{Execution lineage.} Execution lineage records what boundary ran, under what resolved inputs and dependency identities, and what work was materially re-used or re-run.

\paragraph{Artifact lineage.} Artifact lineage records what durable state was produced, consumed, revised, or superseded. It is concerned with maintained state objects rather than prompts or runs alone. This lineage exists both within a step, where intermediate artifacts may depend on one another during local generation, and across steps, where substrate artifacts become explicit dependency edges for downstream execution.

These lineages answer different questions. Prompt lineage answers ``what text led to this?'' Execution lineage answers ``what computation ran?'' Artifact lineage answers ``what durable state exists now, what did it replace, and what downstream state depends on it?''

\subsection{Additive Versus Superseding Semantics}

Agentic systems need a clear distinction between additive and superseding updates.

\paragraph{Additive semantics.} Some steps produce new artifacts without invalidating earlier ones. A new research note, an additional scenario branch, or a fresh evidence extract may coexist with prior artifacts. In additive updates, earlier artifacts remain valid and addressable.

\paragraph{Superseding semantics.} Other steps revise or replace prior artifacts. A current reimbursement summary may supersede a stale version; a revised decision criterion may supersede an earlier criterion; a corrected claim matrix may supersede the earlier matrix it replaces. In these cases, leaving all versions equally active would make the maintained state incoherent.

Supersession is therefore not simply ``another artifact exists.'' It is a semantic relation stating that one artifact replaces another as the current authoritative state for some role and authority scope. The earlier artifact may remain in history for audit, but it should no longer be treated as active current state for downstream consumers.

This distinction is especially important when artifacts are structured payloads rather than undifferentiated prose. Without supersession semantics, downstream consumers may combine incompatible versions. Without additive semantics, the system loses the ability to preserve alternative branches or cumulative notes that were never meant to invalidate prior work.

\subsection{State Quality as a First-Class Evaluation Target}

State quality deserves separate evaluation because a system can repair the visible final answer while still leaving the maintained artifact set incoherent. A final memo may look correct even though an older implementation plan or criteria artifact remains active. Evaluation must therefore test not only whether the final artifact changed appropriately, but whether the maintained artifact set now reflects a coherent current state after revision.

\subsection{Formal Intuition}

Let $A$ be the set of artifacts in a system state. Each artifact $a \in A$ has at least:
\begin{equation}
a = (\iota, r, \tau, p, D, s, L, \alpha)
\end{equation}
where $\iota$ is an identity (addressable/versioned), $r$ is a role, $\tau$ is a type, $p$ is the payload (structured/consumable), $D$ is a dependency set, $s \in \{\mathrm{active},\mathrm{historical},\mathrm{superseded}\}$ is status, $L$ is lineage metadata, and $\alpha$ is an authority scope.

Let $\mathrm{supersedes}(a_{\mathrm{new}}, a_{\mathrm{old}})$ denote that $a_{\mathrm{new}}$ replaces $a_{\mathrm{old}}$ as current state in some role and authority scope. Let $\mathrm{active}(r,\alpha)$ return the authoritative artifact or set of artifacts for role $r$ under scope $\alpha$. Downstream consumers resolve against $\mathrm{active}(r,\alpha)$ unless explicitly pinned to a historical artifact.

An additive step creates a new artifact without changing the active status of existing artifacts in the same authority scope. A superseding step creates a new artifact and updates the status of at least one prior artifact so that downstream consumers know what is now authoritative. Some roles may permit multiple active artifacts, such as alternative scenarios; others may require a single active artifact, such as current decision criteria.

\subsection{Authority Scope and Active-State Resolution}

Authority is not a metaphysical property of an artifact. It is granted by the substrate's resolution policy inside a declared scope. In practice, that policy can be configured by authored graph structure, role rules, human review settings, or runtime governance, but the semantic effect is the same: for a given role and scope, the substrate must be able to determine what counts as current.

For single-authority roles, \texttt{resolve\_active(role, scope)} should return exactly one active artifact. If two artifacts in the same family both claim to be active for the same role and scope, the substrate has a conflict that must be surfaced rather than silently ignored. For multi-authority roles, such as scenario branches, the resolver may return an explicit set of active artifacts together with the branching semantics that permit coexistence.

This yields three cases:
\begin{itemize}
\item \textbf{Single active artifact:} downstream consumers use that artifact by default.
\item \textbf{Multiple active artifacts allowed:} downstream consumers receive the declared active set.
\item \textbf{Multiple active artifacts disallowed:} the substrate enters conflict state and requires explicit repair, override, or branch separation.
\end{itemize}

Authority scope is therefore the resolution boundary that prevents ``current'' from becoming ambiguous. It is what lets one criteria artifact be current for one scenario, another be current for a different scenario, and neither silently override the other outside its declared scope.

\subsection{Lineage Under Revision}

Artifact lineage should therefore record at least four relations:
\begin{itemize}
\item produced-by: which execution step added or revised the artifact in the substrate
\item consumed-by: which downstream boundaries depended on it
\item supersedes: which prior artifacts it replaced, if any
\item superseded-by: which later artifacts displaced it
\end{itemize}

This is the durable-state counterpart to execution-boundary tracking inside the same harness-backed substrate. Artifact lineage tells operators and downstream nodes what maintained state is current, historical, or affected by revision.
\section{Intermediate Artifacts in the Substrate}

\subsection{Execution Boundaries and Maintained Artifact State}

We assume systems that organize work into staged or graph-executed boundaries with explicit upstream and downstream relationships, together with dependency-aware replay under change. Within such systems, this paper defines the artifact-state semantics: what durable artifact state execution boundaries create and maintain, how that state becomes current, and how later revisions change it. An execution step may create many internal artifacts during work and commit one or more artifacts into maintained substrate state. Those committed artifacts, rather than the full local working trace, become the state surfaces later humans and agents consume and improve.

\subsection{Execution Steps and Artifact Boundaries}

Execution-step boundaries are the primary substrate boundaries in the system. A step may internally create evidence digests, tool outputs, criteria notes, claim matrices, partial drafts, synthesis procedures, reconstruction methods, and candidate outputs while it works. Some of these may be persisted for inspection or revision, but only some step-level artifacts become authoritative downstream state in the substrate.

This gives execution steps two responsibilities. First, they localize where maintained-state creation and revision occur. Second, they define the interface contract by which downstream consumers use upstream work. The design tradeoff is granularity: very large steps hide useful intermediate structure, while excessively fine steps create artifact sprawl, schema overhead, and noisy state churn.

\subsection{Granularity Heuristics}

Artifact boundaries should be chosen by expected reuse and revision pressure rather than by trying to materialize every internal thought. Three heuristics are especially useful.

\paragraph{Keep distinct substrate artifacts where downstream dependencies are likely.} If later steps, agents, or humans are likely to consume an intermediate object directly, keep it as its own artifact rather than burying it in a larger step output.

\paragraph{Keep distinct artifacts where review is likely.} If a human reviewer will want to inspect or override a criterion, claim structure, evidence digest, or plan, make that object independently addressable.

\paragraph{Publish where future revision is likely.} If an assumption, constraint, or synthesis rule is likely to change over time, isolate it into an artifact that can be superseded without rewriting unrelated maintained state.

These heuristics suggest several practical defaults. Evidence digests, decision criteria, claim matrices, plans, and synthesis procedures are often good artifact boundaries because they are frequently reused, reviewed, or revised. By contrast, transient formatting drafts, one-off tool logs, and locally redundant decompositions often should remain block-internal unless they later prove to be recurring intervention points.

At a systems level, the question is not ``can this be serialized?'' but ``will keeping this object in the substrate reduce future reconstruction cost more than it increases state-management overhead?'' Good artifact boundaries are therefore the places where revision locality, human oversight, and downstream reuse are most likely to matter.

\subsection{Minimal Reference Architecture}

To make the proposal concrete, consider a minimal substrate with six core parts:
\begin{itemize}
\item a \textbf{graph store} that persists steps, blocks, dependency edges, and authored structure
\item an \textbf{execution harness} that runs the substrate's authored steps and emits step- and block-level artifacts
\item an \textbf{artifact store} that persists payloads, statuses, scopes, and artifact lineage
\item a \textbf{resolver} that answers which artifacts are currently authoritative for a given role and scope
\item a \textbf{supersession manager} that marks replaced artifacts non-active and emits invalidation events
\item an \textbf{editor interface} that lets external humans or agents inspect current artifacts and update steps, blocks, or graph structure
\end{itemize}

One minimal storage model is:

\begin{verbatim}
ArtifactRecord(
  artifact_id,
  family_id,
  role,
  scope,
  status,              # active | superseded | historical
  block_id,
  created_at,
  supersedes_ids[],
  payload_ref,
  payload_hash
)

DependencyEdge(
  src_artifact_id,
  dst_artifact_id,
  edge_type            # consumes | derived_from
)

ActiveIndex(
  role,
  scope,
  artifact_id
)
\end{verbatim}

The storage sketch above mixes core semantics with implementation detail. The core model requires \texttt{artifact\_id}, \texttt{family\_id}, \texttt{role}, \texttt{scope}, \texttt{status}, dependency edges, supersession relations, and payload. Fields such as \texttt{block\_id}, \texttt{created\_at}, \texttt{payload\_ref}, and \texttt{payload\_hash} are useful implementation choices, but they are not logically required by the paper's claim.

\paragraph{Concrete implementation-oriented example.} Consider a scope \texttt{telehealth:baseline}. The active criteria artifact is:

\begin{verbatim}
{
  "artifact_id": "criteria:v1",
  "family_id": "criteria",
  "role": "decision_criteria",
  "scope": "telehealth:baseline",
  "status": "active",
  "depends_on": ["claim_matrix:v2", "tension_analysis:v2"],
  "supersedes": [],
  "payload": {
    "constraint": "Expand access while maintaining feasibility"
  }
}
\end{verbatim}

If a new budget-neutrality requirement arrives, the substrate commits \texttt{criteria:v2} with \texttt{supersedes=["criteria:v1"]}. The resolver then returns \texttt{criteria:v2} as the active artifact for that role and scope, \texttt{criteria:v1} stops resolving as current, downstream dependents such as \texttt{implementation\_plan:v1}, \texttt{recommendation:v1}, and \texttt{final\_memo:v1} are invalidated and regenerated, and unaffected upstream artifacts remain active.

\paragraph{Concrete artifact lifecycle example.} Table~\ref{tab:lifecycle-example} shows the same progression as a state transition sequence.
\begin{table}[H]
\centering
\small
\begin{tabular}{p{0.12\linewidth}p{0.28\linewidth}p{0.25\linewidth}p{0.25\linewidth}}
\toprule
Step & Event & Active state & Regeneration effect \\
\midrule
1 & Commit \texttt{criteria:v1} & \texttt{criteria:v1} active & none \\
2 & Commit \texttt{implementation\_plan:v1}, \texttt{recommendation:v1}, \texttt{final\_memo:v1} & all v1 artifacts active & none \\
3 & Commit \texttt{criteria:v2} superseding \texttt{criteria:v1} & \shortstack[l]{\texttt{criteria:v2} active;\\\texttt{criteria:v1} superseded} & \shortstack[l]{dependents of\\\texttt{criteria:v1} invalidated} \\
4 & Resolve active criteria for downstream steps & \shortstack[l]{\texttt{resolve\_active(}\\\texttt{decision\_criteria,}\\\texttt{telehealth:baseline)}\\$\rightarrow$ \texttt{criteria:v2}} & \shortstack[l]{downstream steps\\consume v2} \\
5 & Regenerate plan, recommendation, memo & \shortstack[l]{\texttt{implementation\_plan:v2},\\\texttt{recommendation:v2},\\\texttt{final\_memo:v2} active} & \shortstack[l]{unaffected upstream\\artifacts preserved} \\
\bottomrule
\end{tabular}
\caption{Concrete artifact lifecycle inside one scope: creation, supersession, active-state resolution, invalidation, and regeneration.}
\label{tab:lifecycle-example}
\end{table}

This reference architecture is intentionally small. It is enough to show the minimal components required for maintained artifact state: the graph of steps and blocks, the execution harness that runs them, the artifact store that preserves identity, family, role, scope, status, dependencies, supersession, and payload, and the resolver/supersession machinery that keeps current state coherent over time. External humans or agents act through the editor interface by inspecting current artifacts and changing steps or structure to improve subsequent regeneration.

What matters here is not only that artifacts are stored, but that they remain regenerable. The harness must know which step produced an artifact, which upstream artifacts were combined to produce it, and which downstream artifacts should be rebuilt when it changes. Generated artifacts are not automatically durable maintained artifacts: stable downstream references require explicit substrate commitment, explicit supersession semantics where relevant, and explicit resolution of what counts as current maintained state.

\subsection{Structured Payloads and Downstream Consumption}

In practice, artifact payloads may be sectioned markdown, XML-like tagged payloads, key-value structures, or JSON-shaped render payloads. The design goal is not maximal formalism. It is preserving enough structure that downstream use and later revision are deliberate rather than heuristic.

\paragraph{Example: decision criteria artifact.}

\begin{verbatim}
{
  "artifact_id": "criteria:v2",
  "role": "decision_criteria",
  "type": "budget_constraint_matrix",
  "status": "active",
  "supersedes": ["criteria:v1"],
  "depends_on": ["utilization_digest:v1", "reimbursement_digest:v1"],
  "payload": {
    "constraint": "Year-one expansion must be budget-neutral",
    "decision_rule": "Prefer phased rollout over full launch",
    "priority_order": ["budget neutrality", "access expansion",
                       "operational feasibility"],
    "open_questions": ["Which sites can absorb added volume?"]
  }
}
\end{verbatim}

The value of these objects is not merely persistence; it is that they provide editable, inspectable, authoritative state boundaries.

\subsection{Worked Example}

Consider a telehealth policy recommendation system organized into staged boundaries:

\begin{verbatim}
source_ingestion
  -> evidence_digest_block
  -> claim_matrix_block
  -> criteria_block
  -> implementation_plan_block
  -> recommendation_synthesis_block
  -> final_memo_block
\end{verbatim}

A set of staged boundaries like this may be realized as a DAG, but the important point here is semantic rather than scheduling-oriented: each boundary creates or revises one or more substrate artifacts while also potentially creating internal intermediates during work.

Downstream blocks consume multiple upstream artifacts simultaneously. For example, a recommendation-synthesis block may resolve:

\begin{verbatim}
recommendation_block consumes:
  - claim_matrix:v2
  - criteria:v3
  - implementation_plan:v1
  - tension_analysis:v2
\end{verbatim}

The synthesis block therefore operates over explicit structured artifacts under known identities rather than over a flattened transcript plus heuristic reconstruction.

\paragraph{Initial state.} The system resolves active upstream artifacts for each step, persists any local intermediate artifacts it wants to inspect, and commits maintained artifacts into the substrate. The initial maintained set might include:

\begin{verbatim}
evidence_digest:v1
claim_matrix:v2
criteria:v1
implementation_plan:v1
recommendation:v1
final_memo:v1
\end{verbatim}

\paragraph{New upstream change.} A new upstream change arrives: a new constraint, a new fact, a new piece of research, or a human-directed shift in direction. In this example, the change is: ``Year-one rollout must be budget neutral.'' The system revises the criteria step, which commits \texttt{criteria:v2} and marks \texttt{criteria:v1} as superseded in the same authority scope.

\paragraph{Affected downstream state.} The system then identifies which downstream maintained artifacts depend on the superseded criteria artifact. In this example:

\begin{verbatim}
criteria:v1 -> superseded by criteria:v2

revised:
  implementation_plan_block
  recommendation_synthesis_block
  final_memo_block

preserved:
  evidence_digest:v1
  tension_analysis:v2
\end{verbatim}

The evidence digest remains valid because the new constraint did not change the upstream evidence basis. The implementation plan, recommendation synthesis, and final memo must be revised because they consume criteria directly or transitively.

\paragraph{Updated maintained state.} The updated substrate state may become:

\begin{verbatim}
criteria:v2            (active; supersedes criteria:v1)
implementation_plan:v2 (active; supersedes implementation_plan:v1)
recommendation:v2      (active; supersedes recommendation:v1)
final_memo:v2          (active; supersedes final_memo:v1)
\end{verbatim}

Downstream consumers now resolve against the active artifacts unless explicitly pinned historically. More importantly, the system knows what to rebuild: the criteria artifact is not just linked to downstream work, but sits on a regeneration path that identifies which compounded artifacts must be recomputed when that criteria changes.

\paragraph{Contrast with transcript- or knowledge-base-centric revision.} A transcript-centric system may still rewrite the memo so that it mentions budget neutrality. A knowledge-base-centric system may also update a note or fact and link it to other notes. What they do not usually provide by default is explicit localization of the revision to the criteria artifact together with scoped downstream recomputation and preservation of unaffected upstream artifacts. The artifact-centric system makes those operations first-class substrate behavior rather than heuristic prompt reconstruction or manual relinking.

\subsection{Cross-Domain Portability}

The telehealth case is only one instance of the abstraction. The same artifact model applies in other domains.

\paragraph{Software engineering.} A coding agent may maintain requirement interpretations, repository findings, migration plans, patch sets, and verification summaries as separate artifacts in the substrate. If a security requirement changes, the system can supersede the requirement artifact and localize recomputation to the patch and test-plan artifacts without discarding unrelated repository analysis.

\paragraph{Legal analysis.} A legal agent may maintain source holdings, jurisdiction filters, issue trees, argument matrices, and draft recommendations in the substrate. If a controlling precedent changes or a jurisdiction is narrowed, the affected authority artifact can be superseded while preserving unrelated fact summaries and issue decomposition artifacts.

\paragraph{Research synthesis.} A literature agent may maintain source extracts, inclusion criteria, evidence tables, uncertainty summaries, and synthesis methods in the substrate. If a study is retracted or a search window expands, the system can supersede the evidence table and regenerate only the dependent synthesis artifacts while preserving unaffected extraction work.

\paragraph{Product planning.} A planning agent may maintain customer-signal summaries, prioritization criteria, dependency maps, roadmap options, and launch plans in the substrate. If executive strategy shifts from growth to retention, the prioritization artifact can be revised directly and downstream roadmap artifacts recomputed without redoing every upstream interview summary.

\section{Evaluation Implications}

Final-output acceptability is insufficient for agentic systems whose intermediate products continue to shape later action. A system can produce an acceptable final answer while leaving stale, contradictory, or unauthoritative intermediate state behind. The relevant evaluation criteria are therefore properties of maintained artifact state.

\begin{table}[H]
\centering
\small
\begin{tabular}{p{0.24\linewidth}p{0.31\linewidth}p{0.33\linewidth}}
\toprule
Criterion & What it detects & Example failure mode \\
\midrule
Traceability of final claims & Whether downstream conclusions can be related to the intermediate artifacts that introduced them & A recommendation appears in the memo, but no claim matrix or criteria artifact can be identified as its source \\
Authority correctness & Whether the system can identify which artifacts are current and authoritative & Two criteria artifacts remain active and downstream consumers resolve them heuristically \\
Supersession correctness & Whether revisions displace the right prior artifacts in the right scope & A corrected claim matrix is created, but the stale matrix remains effectively current \\
Preservation of unaffected artifacts & Whether unrelated state remains stable under local change & Updating a budget constraint rewrites an unrelated access-analysis artifact \\
Stale artifact detection & Whether obsolete or contradictory state is surfaced after revision & The final memo is updated while an older implementation plan still looks current \\
Revision localization & Whether the system supports correction at the intermediate source rather than only at the final surface & A false assumption in an evidence digest can only be addressed by rewriting the memo \\
Preservation of reusable intermediate work & Whether valid partial products, plans, and intermediate structures survive revision & A useful phased rollout plan is discarded because it was never materialized as maintained state \\
Preservation of reasoning- and agent-relevant structures & Whether criteria, transformations, procedures, and similar structures remain available for reuse & A decision rule is flattened into prose and must be reconstructed heuristically later \\
Durable intervention surfaces & Whether the system exposes enough stable points for later improvement & Every revision has to begin from the final answer because no editable upstream artifacts remain \\
\bottomrule
\end{tabular}
\caption{Evaluation criteria for maintained artifact state. These criteria target represented state quality rather than one-shot output quality.}
\label{tab:evaluation-criteria}
\end{table}

These criteria operationalize the maintained-state implications of the data model rather than promising a separate benchmark contribution.

\subsection{From Criteria to Measurable Tests}

The evaluation section becomes more useful when each criterion maps to an executable test over artifact state. A minimal benchmark can therefore define perturbation tasks with known artifact graphs, controlled edits, and gold current-state annotations.

\paragraph{Authority correctness.} Given a task with one gold-active artifact per single-authority role, score whether the runtime resolves the correct artifact after revision. A direct metric is active-resolution accuracy:
\begin{equation}
\mathrm{AuthorityAcc} = \frac{\#\text{roles whose resolved active artifact matches gold}}{\#\text{single-authority roles}}
\end{equation}

\paragraph{Stale artifact detection.} Inject revisions that should supersede known artifacts, then measure whether the system flags stale-but-still-reachable artifacts. This can be scored as stale detection precision, recall, and $F_1$ over artifacts labeled stale in the gold state.

\paragraph{Revision localization.} For each edit, define the gold minimal affected artifact set. Compare the runtime's recompute or rewrite set against that gold set using overreach and miss rates:
\begin{equation}
\mathrm{LocalizationPrecision} = \frac{|R \cap G|}{|R|}, \quad
\mathrm{LocalizationRecall} = \frac{|R \cap G|}{|G|}
\end{equation}
where $R$ is the revised artifact set and $G$ is the gold affected set.

These metrics can be instantiated with perturbation families already natural for agent work:
\begin{itemize}
\item \textbf{Authority-swap edits:} introduce a new criterion or policy summary that should replace a prior one
\item \textbf{Local correction edits:} fix one evidence digest while leaving sibling branches untouched
\item \textbf{Branch-isolated edits:} modify an unrelated branch and test whether unaffected artifacts remain active and unchanged
\item \textbf{Transitive-impact edits:} change an upstream artifact whose dependents span multiple downstream stages
\end{itemize}

\paragraph{Worked benchmark-style example.} Consider a benchmark instance with active artifacts
\{\texttt{claim\_matrix:v2}, \texttt{criteria:v1}, \texttt{implementation\_plan:v1}, \texttt{recommendation:v1}, \texttt{final\_memo:v1}\}
in scope \texttt{telehealth:baseline}. The perturbation is a new budget-neutrality requirement. The gold active-state set after revision is
\{\texttt{claim\_matrix:v2}, \texttt{criteria:v2}, \texttt{implementation\_plan:v2}, \texttt{recommendation:v2}, \texttt{final\_memo:v2}\},
with \texttt{criteria:v1}, \texttt{implementation\_plan:v1}, \texttt{recommendation:v1}, and \texttt{final\_memo:v1} marked superseded. The gold recomputation set is
\{\texttt{criteria}, \texttt{implementation\_plan}, \texttt{recommendation}, \texttt{final\_memo}\},
while \texttt{claim\_matrix} remains unchanged.

If a system instead leaves \texttt{criteria:v1} active, then \texttt{AuthorityAcc}=0 for the \texttt{decision\_criteria} role in that scope. If it recomputes the claim matrix unnecessarily, localization precision drops. If it fails to regenerate \texttt{recommendation} or \texttt{final\_memo}, localization recall drops. If it regenerates the final memo but leaves \texttt{implementation\_plan:v1} active, stale-artifact detection should count that as a miss.

A simple benchmark protocol would report at least three numbers per system: authority resolution accuracy, stale-artifact detection $F_1$, and revision-localization precision/recall. Final-output quality can still be measured separately, but it should no longer stand in for maintained-state quality.

\section{Discussion}

\subsection{Longitudinal Improvability and Reuse}

A system can improve not only by rewriting the final answer, but by editing the intermediate artifact where the weakness originated. An evidence digest can be corrected, a claim matrix can be reorganized, a criterion can be sharpened, a transformation can be revised, or a plan can be recomposed. In many settings, the most valuable output of a model run is not the final prose but an intermediate framing, decomposition, comparison, plan, transformation, reconstruction method, decision procedure, partial product, or unresolved tension that later humans or agents should build upon.

\subsection{Structured Work Requires an Artifact Model}

When work is organized into explicit boundaries with downstream consumers, a state model is needed in addition to execution machinery. Without artifacts as first-class state, structured decomposition remains an execution-time convenience rather than a maintained substrate for revision. The main consequence is practical: a revision can update only the maintained subset of state that actually changed instead of forcing every later improvement to begin again from final prose.

\subsection{Human-AI Collaboration}

Although the paper is not primarily a collaboration paper, the artifact model has clear collaboration implications. Humans can inspect and edit meaningful intermediate boundaries instead of rephrasing high-level instructions into a loop. Other agents can consume the same state with clear expectations about what is current and what has been superseded.

\subsection{Model Failure Modes}

The proposed model introduces its own failure modes, and they should be treated as first-class design risks.

\paragraph{Stale active artifacts.} A revision may correctly produce a new artifact while leaving the prior artifact active in the same role and scope.

\paragraph{Wrong authority resolution.} The resolver may return the wrong active artifact because scope boundaries were modeled incorrectly or because conflict state was silently collapsed.

\paragraph{Schema brittleness.} Overly rigid payload schemas may force unstable work into premature structure, producing artifacts that are formally valid but semantically poor.

\paragraph{Artifact sprawl.} Excessive granularity may create too many maintained artifacts, making review, supersession, and recomputation harder rather than easier.

\paragraph{Final-output/upstream-artifact drift.} A final memo may be regenerated or hand-edited in a way that no longer matches the upstream maintained artifacts that are supposed to justify it.

These are not arguments against the model. They are the main ways an artifact-centric system can fail while still appearing superficially coherent.

\section{Limitations}

The proposal has several important limitations.

\begin{itemize}
\item \textbf{Schema discipline.} Typed artifacts require schema design, substrate rules, and discipline. Poor schemas can create brittle systems or encourage superficial structure.
\item \textbf{Granularity choices.} Artifact boundaries are design decisions. Too coarse, and the system loses update locality. Too fine, and the state model becomes over-fragmented and costly to manage. The heuristics above help, but they do not remove the need for judgment.
\item \textbf{Deliberate substrate entry.} The model depends on deliberate substrate entry. It does not require, and should not encourage, preserving every token of private reasoning or every transient draft.
\item \textbf{Artifact quality.} A persisted artifact can still be incomplete, misleading, or wrong. Durability does not imply truth. A wrong artifact can still be wrong even if it is typed, versioned, and addressable.
\item \textbf{Exploratory fluidity.} Some exploratory tasks may benefit from temporary fluidity before artifact boundaries are chosen. Excessive formalization can reduce flexibility early in the work.
\item \textbf{Stochastic content.} Even when substrate boundaries are explicit, model-generated artifact content remains sensitive to model behavior and runtime conditions. This is a system-level state model, not a claim of universal model determinism.
\item \textbf{Partial abstraction.} Maintained reasoning structures are abstractions over reasoning, not complete records of the model's internal computation. A reconstruction method may itself be wrong, underspecified, or stale.
\item \textbf{Not every task needs it.} Short-horizon tasks, disposable chats, and lightweight ideation may not benefit from durable artifact state.
\item \textbf{Intervention quality.} More intervention surfaces do not guarantee better interventions. Longitudinal improvability is a structural property that still requires useful judgment, validation, and revision practices.
\end{itemize}

The evaluative discussion is also limited. The examples here are intended to clarify what artifact-centric evaluation should measure, not to establish universal performance claims across agentic systems.

\section{Conclusion}

This paper proposes a data model for maintained intermediate work inside a single harness-backed substrate that includes the graph of steps and blocks, the runtime that executes them, the intermediate and final artifacts they generate, and the lineage and currentness/supersession machinery that keeps that state coherent over time. The central claim is not that intermediate artifacts make models smarter. It is that they make AI work maintainable by preserving the state a system needs to inspect, trace, strengthen, supersede, and extend. Without that state, final artifacts become opaque revision targets. With it, revisions can operate on the intermediate state where evidence, criteria, plans, and synthesis procedures were introduced, and the substrate can remain coherent across later human and agent iterations.

\end{document}